\documentclass[letterpaper, 10 pt, conference]{ieeeconf}  

\IEEEoverridecommandlockouts                              

\usepackage{amsmath} 
\usepackage[ruled, linesnumbered]{algorithm2e}
\usepackage{graphicx}
\newcommand{\argmax}{\mathop{\mathrm{argmax}}}
\newcommand{\argmin}{\mathop{\mathrm{argmin}}}   
\newcommand{\euler}{e}

\title{\LARGE \bf
Non-myopic Planetary Exploration Combining In Situ and Remote Measurements
}
\author{Suhit Kodgule, Alberto Candela and David Wettergreen
\thanks{}
\thanks{\textcopyright 20xx IEEE.  Personal use of this material is permitted.  Permission from IEEE must be obtained for all other uses, in any current or future media, including reprinting/republishing this material for advertising or promotional purposes, creating new collective works, for resale or redistribution to servers or lists, or reuse of any copyrighted component of this work in other works.}
}
\begin{document}
\maketitle
\thispagestyle{empty}
\pagestyle{empty}
\begin{abstract}

Remote sensing can provide crucial information for planetary rovers. However, they must validate these orbital observations with in situ measurements. Typically, this involves validating hyperspectral data using a spectrometer on-board the field robot. In order to achieve this, the robot must visit sampling locations that jointly improve a model of the environment while satisfying sampling constraints. However, current planners follow sub-optimal greedy strategies that are not scalable to larger regions. We demonstrate how the problem can be effectively defined in an MDP framework and propose a planning algorithm based on Monte Carlo Tree Search, which is devoid of the common drawbacks of existing planners and also provides superior performance. We evaluate our approach using hyperspectral imagery of a well-studied geologic site in Cuprite, Nevada.  

\end{abstract}

\section{INTRODUCTION}
\begin{figure}
    \centering
    \includegraphics[width=8.4cm]{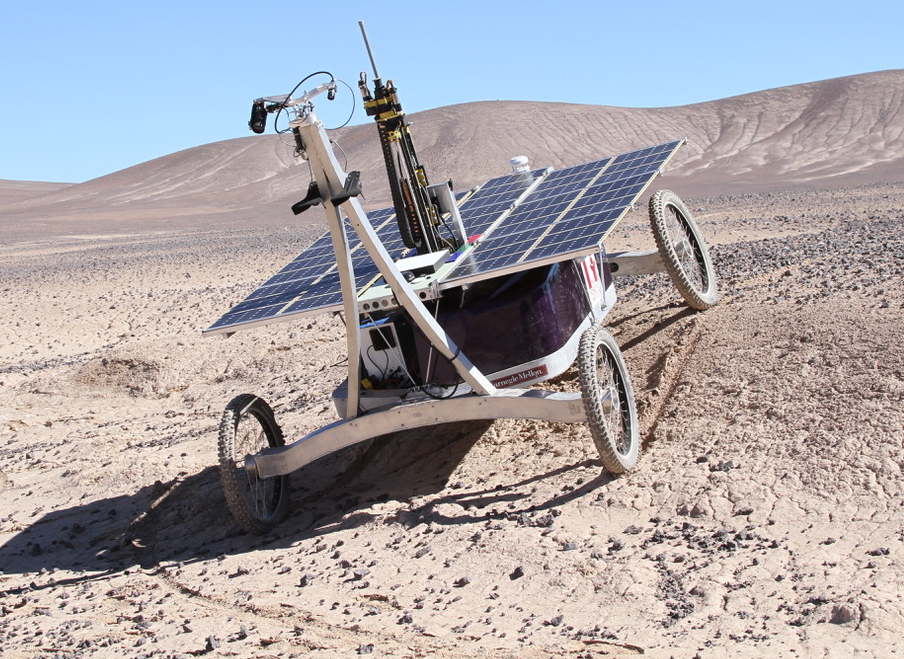}
    \caption{Prototype planetary rover in the Atacama Desert of Chile, exploring terrain to classify and map geology.  An on-board spectrometer measures individual rocks to identify mineralogical composition.}
    \label{zoe}
\end{figure}

An emerging field of interest in space exploration is improving the autonomy of planetary rovers that are utilized for remote exploration and survey. Figure \ref{zoe} displays a prototypical rover utilized for conducting field experiments.  These robots must be capable of effectively incorporating science data into their planning strategies for satisfying mission objectives. Typical objectives include measuring discrete features of interest \cite{estlin2012aegis} or informative path planning based on spatial experimental design \cite{doi:10.1002/rob.20391}.   

Remote sensing measurements play an integral role in not only meeting these objectives, but also prescribing new objectives for science-aware exploration. For instance, spectroscopy \cite{GREEN1998227} \cite{PLAZA2009S110}, that is heavily reliant on orbital sensor measurements is used extensively for understanding mineralogical composition on planetary surfaces. Ensuring that these measurements are accurate and provide an unambiguous interpretation of the surface then is critical for guaranteeing the success of the mission. However, orbital hyperspectral sensors have poor resolution and as a result, features that are concentrated on a small region can go undetected. \textit{In situ} sensing can aid in detecting these exceptions as they possess a higher resolution and also provide a higher signal to noise ratio.

This work focuses on the \textit{spatio-spectral} exploration (SSE) problem \cite{Thompson-2015-17189}. In this scenario, a field robot gathers a library of \textit{in situ} spectral measurements that facilitates \textit{spectral unmixing} of the remote image. Concretely, it attempts to collect pure spectral signatures that could reconstruct the orbital image. A detailed explanation of SSE is provided in Section \ref{bg}.  

In this scenario, a path planning algorithm must collect a set of most informative samples while satisfying a sampling constraint. This task bears similarity to the Informative Path Planning (IPP) problem which requires an efficient trade-off between exploration and exploitation. A number of active learning strategies have been employed for IPP scenarios such as search and rescue \cite{singh2009nonmyopic}, sensor placement \cite{krause2008near} or autonomous underwater navigation \cite{5509714}. All of these approaches however, rely on mapping a scalar field representing the environment, which is commonly done using a Gaussian Process (GP). On the other hand, in our scenario the objective function is dependent on highly correlated distributions of minerals which cannot be mapped to a scalar field. Moreover, GP models are unsuitable for solving problems with non-stationary environments. We explain how SSE involves navigating a non-stationary environment in Section \ref{approach}. 
A number of methods have been employed for science-aware navigation. Thompson {\it et al.} developed a GP model using remote sensing observations to predict the content of basalt on the surface \cite{thompson2011autonomous}. Smith designed a POMDP based planner for detecting life in the Atacama desert \cite{smith2007probabilistic}. Woods {\it et al.} designed an autonomous, opportunistic science system which included a replanning strategy for arranging a series of science activities \cite{woods2009autonomous}. Approaches employed for SSE however, continue to utilize naive strategies such as an exhaustive coverage of the region or the Greedy Spectral Selection (GSS) algorithm \cite{Thompson-2015-17189}: a greedy strategy that attempts to maximize the entropy of the collected samples.

Reinforcement Learning (RL) provides a suitable framework for developing non-myopic plans while trading off exploration against exploitation. These properties make RL well suited for generating plans for SSE. Several Dynamic Programming (DP) approaches such as Value Iteration \cite{Bellman} and Policy Iteration \cite{howard:dp} provide provable convergence guarantee. Nevertheless, they require complete knowledge of states and transitions of the environment, which cannot be easily determined in real-world scenarios. In such situations, online Monte-Carlo based approaches prove useful. Morere {\it et al.} use POMDPs with Gaussian Process based reward functions for monitoring spatial phenomena with a UAV \cite{Morere}. Clary {\it et al.} use a Monte Carlo Tree Search (MCTS) with Upper Confidence Bounds for Trees (UCT) for legged locomotion \cite{ICAPS1817789}. Ross {\it et al.} propose a no-regret algorithm for learning non-stationary policies using Imitation Learning \cite{Ross2011ARO}. However, acquiring relevant demonstrations for robotic exploration is a non-trivial problem. 

In this work, we focus on adapting MCTS for SSE. While adapting MCTS, it is important to note that there is inherent uncertainty present in the environment as we do not know the \textit{in situ} spectral information for each state of the environment {\it a priori}. In Section \ref{approach} we demonstrate how utilizing orbital sensing information itself can account for this uncertainty. This paper makes the following contributions:
\begin{itemize}
    \item We propose an MDP framework in non-stationary environments for SSE that uses orbital sensing measurements to account for uncertainty in the environment.
    \item We present a path planning algorithm based on MCTS to collect the most informative samples for spectral unmixing. 
    \item We provide a detailed comparative analysis of our proposed method against current practices in SSE on a spectroscopic dataset obtained from the canonical mineral site in Cuprite, Nevada. 
\end{itemize}
The paper is organized as follows. In Section \ref{bg} we briefly summarize MDP, SSE and Spectral Mixture Models. In Section \ref{approach}, we explain our approach based on Monte Carlo planning that can be utilized to gain informative spectral samples. In Section \ref{exp}, we describe the spectroscopic dataset utilized and the planning strategies that we compare against. Finally we provide our results and concluding thoughts in Sections \ref{res} and \ref{con} respectively.

\section{BACKGROUND}
\label{bg}
We begin by defining an MDP and then provide brief descriptions of spectral mixture models and SSE.
\subsection{Markov Decision Process}
An MDP can be represented by the tuple $M = (S, A, T, R)$. 
\begin{itemize}
    \item $S$ is a finite set of states.
    \item $A$ is a finite set of actions.
    \item $T$ is a probabilistic transition function $T(s,a,s') = p(s' | s, a)$ which defines the probability of transitioning into state $s'$ after taking action $a$ from state $s$.
    \item $R: S\times A \rightarrow \rm I \!R $ is the reward function.
\end{itemize}
MDPs rely on the Markov assumption which states the distribution on future states is only dependent on the current state. 
A \textit{policy} $\pi: S \rightarrow A \, $  is a function that maps  a state $s \in S$ to an action $a \in A$. An optimal policy  $\pi^*$ is defined as the one that maximizes the cumulative discounted rewards. Formally, 
\begin{equation}
    \pi^* = \argmax_{\pi} E \left [ \sum_{t = 0}^{\infty} \gamma^t r_t^{\pi}\right] \label{optimal_policy}
\end{equation}
Here, $\gamma$ is the discount factor and $r_t^\pi$ is the reward acquired at time $t$ by following the policy $\pi$. A detailed explanation of MDPs can be found in \cite{sutton2018reinforcement}.
\subsection{Spectral Mixture Models}
Orbital hyperspectral sensors have poor spatial resolution on account of measuring from high altitudes. As a consequence, these sensors record scenes in which numerous distinct minerals contribute to the same spectrum that is measured from a single pixel of the hyperspectral image. This gives rise to mixed pixels, which is formed by a combination of pure minerals called \textit{endmembers}. The fractions in which these endmembers combine to form the mixed pixel are known as abundances. We consider here a linear mixing model (LMM) which assumes that the mixed pixel's spectrum is formed by a linear combination of the endmembers that constitute the pixel. Formally, if a spectrum has $d$ bands or channels, we can define a pixel of an orbital image by a vector $x \in \rm I \!R^d$. Further, if there are $K$ endmembers observed in the entire orbital image, then each pixel $x_j$ of an orbital image $X$ can be expressed as:
\begin{align}
    x_j = \sum_{i=1}^K a_{ij} y_i \quad \forall x_j \in X 
    \label{unmixeq}
\end{align}
where $y_i \in \rm I \!R^d$ is the $i_{th}$ endmember spectrum and $a_{ij}$ is the abundance scalar for $i_{th}$ endmember and $j_{th}$ pixel. For physical realizability, endmembers must not have negative abundances, that is: $a_{ij} \geq 0, \ \forall i,j$ and $\sum_{i=1}^K a_{ij} = 1, \ \forall j$. Equation (\ref{unmixeq}) can be written as $x_j = Y a_j, \ \forall x_j \in X$, where $Y$ is a library of endmember spectra.

An unmixing problem attempts to find $a_j$ given $Y$ and $X$ for all pixels of $X$. Typically, this is solved using non-negative least squares optimization \cite{trove.nla.gov.au/work/21391104} and involves solving the following objective for each pixel:
\begin{align}
    \argmin_{a_j} \quad \lVert Y a_j - x_j \rVert_2 \\
    s.t. \ a_j \geq 0 \nonumber
\end{align}
\subsection{Spatio-Spectral Exploration}
SSE involves collecting a library of endmembers $Y$ that can solve an unmixing problem. Consider a robot that collects $N$ {\it in situ} spectral measurements along a path. Let the set of sampling locations along the path be $B$. Further, let the \textit{in situ} spectral measurements be defined as  $y = f(b) + \epsilon, \ \forall b \in B$, where $\epsilon$ is a sensor noise sampled from a normal distribution: $\epsilon \sim \mathcal{N}(0, \sigma)$. The set of all {\it in situ} measurements on path $B$ is then  $Y_B = \{y_i : y_i \in \rm I \!R^d , \ 1 \leq i \leq N\}$. Formally, the robot must collect samples that minimize the objective:
\begin{equation}
    O(B) = E\left [\sum_{x_j \in X} \min_{a_j} \left \Vert Y_B a_j - x_j \right \Vert_2 \right]
    \label{risk}
\end{equation}
\begin{displaymath}
s.t. \ a_j \geq 0, \  C(B) \leq \beta
\end{displaymath}
where $C(B)$ is the sampling cost for the rover.

In its present form, (\ref{risk}) cannot be solved during the traverse as all the elements of $Y_B$ are not available. However, the solution can be approximated by replacing the elements in $Y_B$ of unvisited locations by the corresponding remote sensing measurements. Let $V_t \subseteq B$ be the set of locations the rover has sampled until time $t$ and $Y_{V_t}$ be the corresponding {\it in situ} spectra. Further, let $D_t \subseteq B$ be the set of unvisited locations and $X_{D_t}$ be the set of remote sensing measurements at those locations. Then as demonstrated in \cite{Thompson-2015-17189}, solving (\ref{risk}) is analogous to solving the following objective function:
\begin{equation}
    O(B|V_t) = \sum_{x_j \in X} \min_{a_j} \left \Vert [Y_{V_t} \; X_{D_t}] a_j - x \right \Vert_2 
    \label{risk_app}
\end{equation}
\begin{displaymath}
 s.t. \ a_j \geq 0, \  C(V_t) + C(D_t) \leq \beta
\end{displaymath}

\section{Methodology}
\label{approach}
The objective of our planner is to provide sampling locations that satisfy Equation (\ref{risk_app}). For this purpose, we first describe our planning environment and propose an MDP for navigating in that environment. 
\subsection{MDP Formulation}
We first address the inherent uncertainty present in the environment on account of \textit{in situ} sensing. As the future \textit{in situ} spectra will not be available at present time, this makes our environment partially observable. While it is certainly possible to apply POMDP based techniques to this problem, we display how using the inherent structure of SSE allows us to reduce this problem to solving multiple MDPs with complete observability. We now describe this MDP. 
\subsubsection{State Definition}
Formally, we define the state $s$ as a tuple $(Y, V, X_{D}, D, X)  $ where $V, D$ are sets of visited and unvisited locations respectively, $Y$ is the set of {\it in situ} spectra collected from the visited locations, $X_{D}$ is the set of remote observations corresponding to the locations in $D$ and $X$ is the orbital image. This definition is necessary to hold the Markov assumption. However, a consequence of including previous measurements into the state definition is that the size of the state-space increases significantly. In this regard, utilizing MCTS for finding the optimal policy is beneficial as the branching factor for the tree is only dependent on the size of the action-space. We limit our action-space to 8 actions to maintain tractability.  
\subsubsection{Action Definition}
Let the set of all actions be $A$. As we do not have access to \textit{in situ} measurements, we consider an action as adding a remote observation measurement to the spectral library. Thus, an action in this MDP corresponds to adding a new location to $D$ and the corresponding remote sensing observation to $X_{D}$. An action is said to be valid if the new location is reachable from the current rover's position. We assume the rover can only traverse on an eight-connected grid and the reachable locations are the immediate neighboring locations on the grid. For verbosity, let the set of valid actions for a state $s$ be $\Omega(s)$. The transition function $T$, is deterministic as every action from a particular state has a unique successor. 

Notice that in this definition of MDP, the number of \textit{in situ} measurements remains fixed for a particular MDP. Thus, at each time step $t$, all the states for $MDP(t)$ possess only those \textit{in situ} measurements collected till time $t$, where $MDP(t)$ refers to the MDP generated at time $t$. This ensures full observability. While we cannot  utilize Equation (\ref{risk}) for solving this MDP, the simplification described in Equation (\ref{risk_app}) can be fully defined using the state and actions as described by our MDP. This then allows us to develop planning strategies while maintaining tractability.

\subsubsection{Reward Definition}
Computing NNLS error and LS error for a given library of spectra is computationally expensive on account of the matrix inversions required for computing its solution. Consequently, it is unfeasible to define our reward function as either of the two metrics. We chose to use the differential entropy as a reward for solving this MDP. Differential entropy has a closed form solution and is independent of the size of the orbital image. Moreover, differential entropy has been previously applied for science-aware exploration and shown to provide competent results \cite{Thompson-2008-10064,Gautam8206232}. A brief description of it is given below.

Let $X$ be a random variable with a probability density function $f$ whose support is $\chi$. Then the differential entropy is defined as \cite{cover2006elements}:
\begin{equation}
    h(X) = -\int_{\chi} f(x)log\left ( f(x) \right ) dx.
    \label{diffe}
\end{equation}
For a multivariate Gaussian random variable with covariance $\Sigma$, (\ref{diffe}) has the following closed form solution \cite{30996}:
\begin{equation}
   h(X) = \frac{1}{2} \ln{|2\pi \euler \Sigma|}.
   \label{mvgentropy}
\end{equation}
For a state $s$ in our MDP, we define our reward function $R(s)$ as:
\begin{equation}
\label{reward}
    R(s) = h(S) = \frac{1}{2} \ln{|2 \pi \euler \Sigma_{S,S}|} - \tau U(S).
\end{equation}
where $S = [Y, \ X_{D}]$ and $U(S)$ is a function that penalizes paths with multiple visits to the same locations. 
\subsection{Non-myopic Planner for Spatio-Spectral Exploration}
Here, we describe our planner named Non-Myopic Planner for Spatio-Spectral Exploration (NMPSE). Our approach comprises of creating and solving multiple trees using MCTS in a sequential manner for each new MDP generated. The algorithm is shown in Algorithm \ref{MCTSALGO}. First, the current state is passed to the MCTS solver. The algorithm constructs a tree with the root as the current state and returns the optimal action for that state. The optimal action for this MDP consists of a sampling location on the map. Next, an \textit{in situ} spectral measurement is taken at the location alluded by the optimal action. A new state is created by appending the new location to the set of visited locations and the sampled \textit{in situ} spectral measurement to the set of {\it in situ} measurements. This is represented by the $Traverse$ function in the algorithm For this new state, we again construct a tree employing MCTS with the new state as the root node and find the next optimal action. In this way, additional \textit{in situ} samples are collected until the sampling budget is exhausted. We now explain MCTS as applied to SSE.

MCTS is a partial tree search algorithm that  efficiently balances the exploration versus exploitation trade-off. Each node in the tree corresponds to a particular state $s$ of the MDP. Applying an action $a$ from $s$ results in a successor state $s'$. This is depicted in the $Step$ function. The function returns $s'$ and a reward $r'$ which is computed according to Equation (\ref{reward}). The algorithm only expands nodes that result from applying an action $b \in \Omega (s)$. 

If an action of the current node has not been tried before, that action is given precedence. The algorithm then rolls out from the resulting successor state until a fixed depth and returns the cumulative discounted reward. On the other hand, if all feasible actions of a state have been tried, the next action to take is based on the Upper Confidence Bound metric that efficiently trades-off exploration against exploitation. Line (\ref{eve}) in Algorithm \ref{MCTSALGO} shows this specific metric. This process is repeated for the successor state until a fixed depth is reached. 

The first new successor state resulting from either of the two cases is appended to the tree and the reward is backed up to the root of the tree. Finally, the action leading to the successor with the maximum cumulative reward from the root is returned. Note that all the nodes of the tree constructed from a particular MCTS call will have the same set of visited locations as no action in the MDP causes the rover to physically move to another location.

A key distinction between our method and traditional MCTS approaches is we use a particular MCTS tree to find just one optimal action. Once an optimal action has been found, the agent executes the action and constructs a new MCTS tree from the successor state. This is necessary as the previous tree does not possess information about the additional {\it in situ} measurement.
\begin{algorithm}
\SetAlgoNoLine
\DontPrintSemicolon
\SetKwFunction{FNMPSE}{NMPSE}
  \SetKwProg{Pn}{function}{}{}
  \Pn{\FNMPSE{$v_0, \ y_0$}}{
        $V \leftarrow v_0$, $Y \leftarrow y_0$ \;
        $X_D, D \leftarrow \emptyset$ \;
        $s \leftarrow \{Y, V, X_D, D, X \} $\;
        \While{$\lvert Y \rvert < \beta$}{
            $T(root) \leftarrow s$\;
            $a^* \leftarrow$ MCTS($T$)\;
            $v_{new},$ $y_{new} \leftarrow$ Traverse($a^*$)\; 
            $V \leftarrow \{V \cup v_{new} \}$\;
            $Y \leftarrow \{Y \cup y_{new} \}$\;
            $X_D, D \leftarrow \emptyset$ \;
            $s \leftarrow \{Y, V, X_D, D, X\}$\;
        }
 }
\SetKwFunction{FMCTS}{MCTS}
  \SetKwProg{Pn}{function}{}{}
  \Pn{\FMCTS{$T$}}{
        \For{$i=0$ \KwTo max\_iterations}
        {
           Simulate($T(root)$, 0)\; 
        }
        \KwRet $\argmax\limits_{a} Q(s_0,a)$\;
  }
\SetKwFunction{FSimulate}{Simulate}
  \SetKwProg{Fn}{function}{}{}
  \Fn{\FSimulate{$s$, $depth$}}{
        \If{$\gamma^{depth} < \epsilon$}{
            \KwRet 0
        }
        \If{$s$ has untried actions}
        {   Sample $a$ from untried actions\;
            $(s',r) \leftarrow$ Step($s$,$a$)\;
            \KwRet $r + \gamma \cdot$Rollout($s'$, $depth+1$)\;
        }
        $a \leftarrow \argmax\limits_{b \in \Omega(s)} Q(s,b) + \kappa \sqrt{\frac{log(N)}{N_a}}$ \label{eve}\;
        $(s',r') \leftarrow Step(s,a)$\;
        $G \leftarrow r' + \gamma \cdot$Simulate($s',depth + 1$)\;
        $N(s) \leftarrow N(s) + 1$\;
        $N(sa) \leftarrow N(sa) + 1$\;
        $Q(s,a) \leftarrow Q(s,a) + \frac{G - Q(s,a)}{N(sa)}$\;
        \KwRet G
  }
\SetKwFunction{FRollout}{Rollout}
  \SetKwProg{Fn}{function}{}{}
  \Fn{\FRollout{$s$, $depth$}}
  {
    \If{$\gamma^{depth} < \epsilon $}{\KwRet 0} 
    $a \sim \Omega(s)$\;
    $(s',r') \leftarrow$ Step($s$, $a$)\;
    \KwRet $r' + \gamma \cdot$Rollout($s'$, $depth + 1$)\;
    
  }
\SetKwFunction{FStep}{Step}
  \SetKwProg{Fn}{function}{}{}
  \Fn{\FStep{$s$, $a$}}
  {
    $\{Y,V, X_D, D, X \} \leftarrow$ Unwrap(s)\;
    $d_{new}, x_{new} \leftarrow$ Unwrap(a)\;
    $s' \leftarrow \{ Y, V, \{X_D \cup x_{new} \}, \{D \cup d_{new} \} , X \}$\;
    $r' \leftarrow Reward(s')$ \;
    \KwRet $(s', \ r')$\;
    
  }
  \caption{NMPSE}
  \label{MCTSALGO}
\label{MCTS algo}
\end{algorithm}
\begin{figure}
    \centering
    \includegraphics[width=8.4cm]{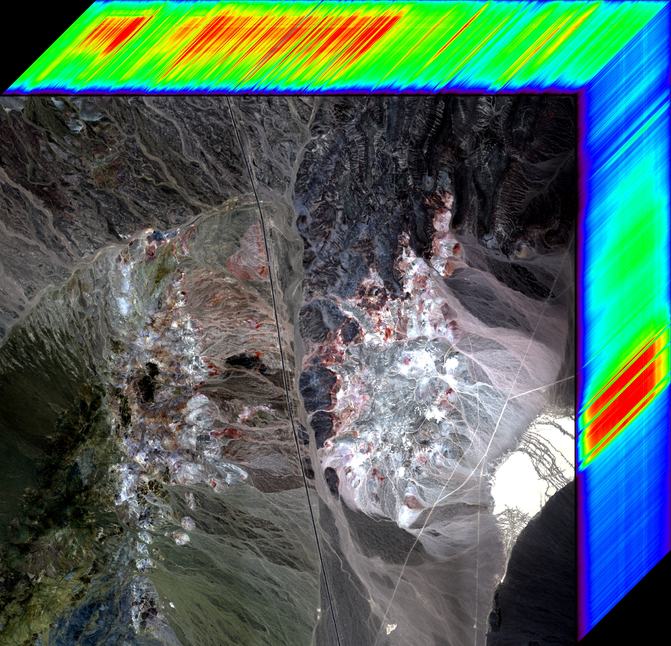}
    \caption{Spectroscopic map of Cuprite as measured by AVIRIS-NG.}
    \label{fig:cuprite}
\end{figure}
\section{Experiments}
\label{exp}
\subsection{Environment Description}
We simulate an exploration scenario using a high resolution data acquisition system. Specifically, we use the data obtained by Airborne Visible Near Infrared Spectrometer - New Generation (AVIRIS-NG) \cite{GREEN1998227,HAMLIN5747395} as  proxy for {\it in situ} spectra. AVIRIS-NG observes spectra in the range of $[0.38\mu m, \ 2.5\mu m]$ at a spectral resolution of $5 nm$ and an exceptionally high spatial resolution (3.7 m/pixel). Studies have shown that AVIRIS-NG measurements are a good analog to {\it in situ} infrared spectra \cite{Thompson2018a}, with inferences drawn on AVIRIS-NG transferring well onto data collected by \textit{in situ} sensing instruments.

For remote sensing measurements, we use data obtained by the Advanced Spaceborne Thermal Emission and Reflection Radiometer (ASTER) \cite{Fujisada}. The instrument consists of multiple cameras possessing three visible short-wave infrared (VSWIR) and six short-wave infrared (SWIR) bands. In fact, other studies have also used AVIRIS-NG and ASTER measurements as proxies for low and high resolution spectral measurements, respectively \cite{Thompson-2015-17189,Kruse2007,Thompson2018b}. We evaluate our approach on measurements taken at from a mining district in Cuprite, Nevada; a well-studied site with high mineralogical diversity \cite{Swayze2014}. Figure \ref{fig:cuprite} shows an example of the AVIRIS-NG spectroscopic map of Cuprite. We were able to associate the two instruments' observations by aligning them with respect to both their spatial and spectral dimensions. We first registered both images with a planar homography approach. We then used the empirical line method \cite{Smith1999} to find the correspondence between the ASTER and AVIRIS-NG reflectance values. 
\subsection{Comparison with Baselines}
Next, we empirically evaluate out approaches against a pair of uninformed planners: Fixed Step and Random, and an informed planner: GSS. A brief description of them is provided below.
\subsubsection{Fixed Step} This strategy sequentially samples waypoints uniformly along one direction until the sampling budget is exhausted.
\subsubsection{Random} In this case, the robot randomly samples a location from the reachable locations of the robot.
\subsubsection{Greedy Spectrum Selection} We implemented the static planner named Greedy Spectrum Selection (GSS). The algorithm is displayed in Algorithm \ref{gss_algo}. For a detailed explanation, refer \cite{Thompson-2015-17189}. The algorithm has been modified to use the MES loss instead of NNLS loss. It greedily selects the next sample from a set of waypoints that maximizes (\ref{mvgentropy}) under some traversal constraint. It does not adaptively replan the path after each {\it in situ} spectral measurement is collected. In order to conduct a fair comparison between the two, we provide the same start locations for both the algorithms. Further, we define the waypoints as all the points on the eight-connected grid which our rover can traverse in the MDP. In addition, GSS algorithm requires a goal location and a path traversal budget. We pass the same goal location and the traversal cost observed during NMPSE evaluation for a particular initial rover configuration. 
\begin{algorithm}
\label{gss_algo}
\SetAlgoNoLine
\DontPrintSemicolon
\SetKwFunction{FGSS}{GSS}
  \SetKwProg{Pn}{function}{}{}
  \Pn{\FGSS{$V_{start}$, $V_{end}$, $X$, $\beta$, $W$}}{
        $Q \leftarrow \{V_{start}, V_{end}\}$ \;
        \While{$C(Q) < \beta$}{
            \ForAll{$v \in W \setminus Q$}{
                $X_V \leftarrow$  $\{X_Q,X_v\}$\;
                $R(Q \cup v) = \frac{1}{2}\ln{|2 \pi \euler \Sigma_{X_V, X_V}|}$\;
                \If{$R( Q \cup v) > R^*$}{
                    $R^* \leftarrow R(Q \cup v)$\; 
                    $v^* \leftarrow v$
                }
            }
            $ Q \leftarrow \{ Q, v^* \}$\;
        }
 }
 \caption{Greedy Spectral Selection}
\end{algorithm}

\noindent We used mean reconstruction error (MRE) of the collected samples as the performance metric in all of our evaluations. The metric is simply (\ref{risk}) averaged over 50 trial runs.
Apart from comparing NMPSE with the planners mentioned above, we also evaluate the effects of varying the hyperparameters of the MCTS in NMPSE on its computational efficiency.

\section{Results}
\label{res}
\subsection{Evaluation of MCTS Parameters}
\begin{table}[]
\caption{Comparison of Mean Reconstruction Error (RE) and average computation time for MCTS with different maximum tree depths.}
\label{dptable}
\centering
\begin{tabular}{|l|l|l|}
\hline
Depth & Mean RE       & Average Action \\
      & NMPSE       & Time (sec)     \\ \hline
5     & 783.64      & \textbf{0.7451}        \\ \hline
7     & 776.1989 & 1.0570            \\ \hline
10    & \textbf{758.3488} & 1.5304            \\ \hline
\end{tabular}
\end{table}
We first evaluate the performance of NMPSE by varying the depth of the MCTS. Increasing the depth of the tree would cause the planner to consider longer horizons in its planning strategy. This in turn, would lead to an increase in the computation cost. The discount factor $\gamma$ was kept at 0.9 and the number of Monte-Carlo simulations during each MCTS call were kept at 500. MRE and the average action time which is the mean time taken by MCTS to provide one optimal action is provided in Table \ref{dptable}. The MRE decreased with increasing depth. However, the average action time increased at a faster rate. A depth of 5 seemed ideal for this problem as it allowed us to keep the computation time at less that one second. All the experiments explained below are implemented with 500 Monte-Carlo simulations, a max depth of 5 and a discount factor of 0.9.
\subsection{Evaluation of NMPSE}
We divided the evaluation of NMPSE into two parts. The first evaluation was done by constraining the number of samples the rover can collect and the second evaluation was conducted by constraining the path length. A detailed explanation of each of the evaluations is given below. 
\subsubsection{Constraining Path Length}
\begin{figure}
    \centering
    \includegraphics[width=8.4cm]{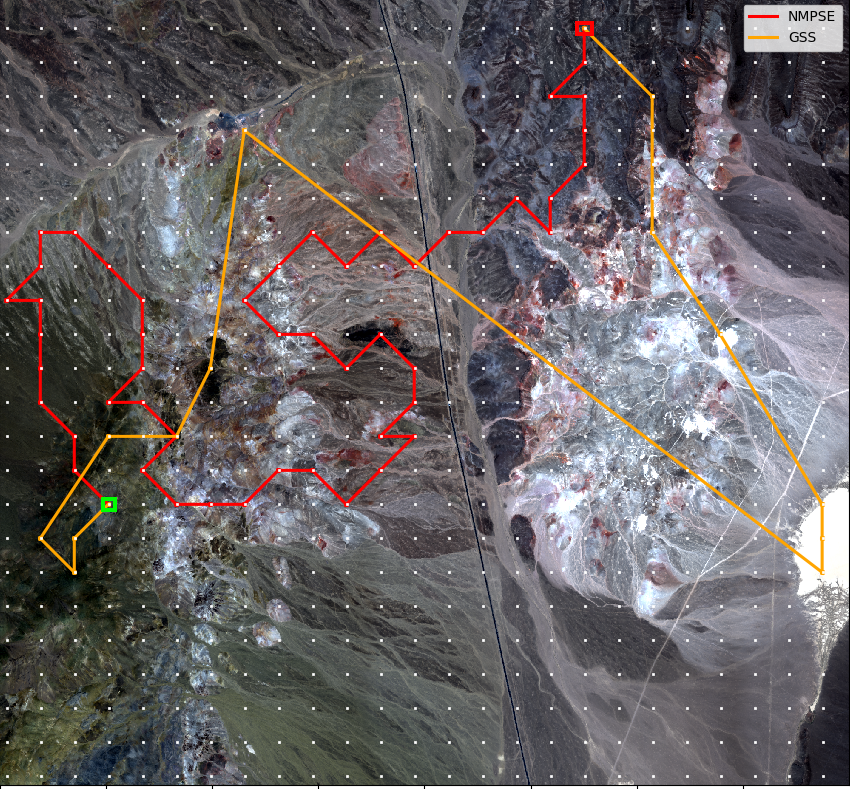}
    \caption{Example paths from NMPSE (red) and GSS (orange) when constrained by traversal budget overlaid on the map of Cuprite mining site. The white dots on image refer to discretized grid of sampling points. The green and red squares correspond to start and end locations of the traverse.}
    \label{gss_paths}
\end{figure}
\begin{table}[]
\caption{Comparison of Mean Reconstruction Error (MRE) and t-test for NMPSE and GSS}
\label{SMPTable}
\centering
\begin{tabular}{|l|l|l|l|}
\hline
Path   & MRE       & MRE      & $p < 0.05$\\ 
Length & NMPSE            & GSS             &  \\ \hline
250    & \textbf{1200.79} & 2533.62         & Y            \\ \hline
500    & \textbf{783.64}  & 2033.13         & Y            \\ \hline
750    & \textbf{592.19}  & 1066.78         & Y            \\ \hline
1000   & \textbf{469.61}  & 563.38          & Y            \\ \hline
1250   & 396.32           & \textbf{372.53} & \textbf{N}   \\ \hline
\end{tabular}
\end{table}
As GSS is constrained by traversal cost instead of sampling cost, it is more equitable to compare the performance of the planners against path length. The path cost is defined as 10 units for traversing to one of the neighbors of a point on the eight-connected grid. In a real-world scenario, this is equivalent to traversing approximately 300m at the Cuprite Site. NMPSE outperforms GSS in all but one case. Surprisingly, GSS is able to achieve comparable performance only with a traversal cost of 1000 and above. It was expected that GSS would more efficiently choose waypoints with a smaller path budget. One explanation for this discrepancy is as GSS is myopic in nature, it finds it difficult to choose additional waypoints that do not exceed the traversal cost with a smaller path budget. Figure \ref{gss_paths} demonstrates this behavior. GSS greedily samples the most informative location, despite it requiring a significant portion of the traversal budget. This leaves it with a small portion of the budget in the latter parts of the traverse.

\begin{figure}
    \centering
    \includegraphics[width = 8.2cm]{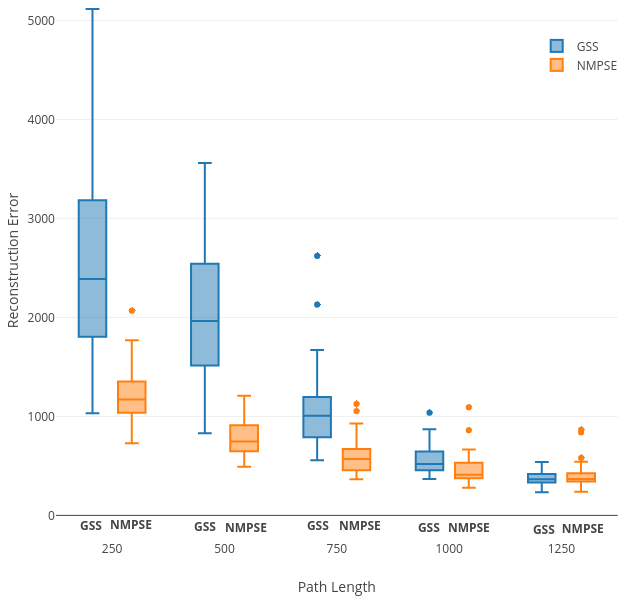}
    \caption{Comparison of the performance of NMPSE (orange) against GSS (blue) with varying path lengths. Both planners were initialized with the same rover configurations.}
    \label{SNMvsNMPSE}
\end{figure}

Figure \ref{SNMvsNMPSE} displays the corresponding boxplot. MRE for both the approaches and whether statistically significant difference was observed is displayed in Table \ref{SMPTable}. We evaluated statistical significance with a  one-tailed t-test: comparing NMPSE with GSS. While GSS achieved better performance than NMPSE for path length of 1250, it did not achieve statistical significance. Moreover, from a practical point of view, rover traverses rarely exceed more than 1-2 km in a day \cite{wettergreen2005second} which roughly translates to around 750 units of  path length in our simulated experiment. NMPSE then would provide better results compared to GSS in field experiments.
\subsubsection{Constraining Sampling Budget}
\begin{figure*}[t]
    \centering
    \includegraphics[width=16cm]{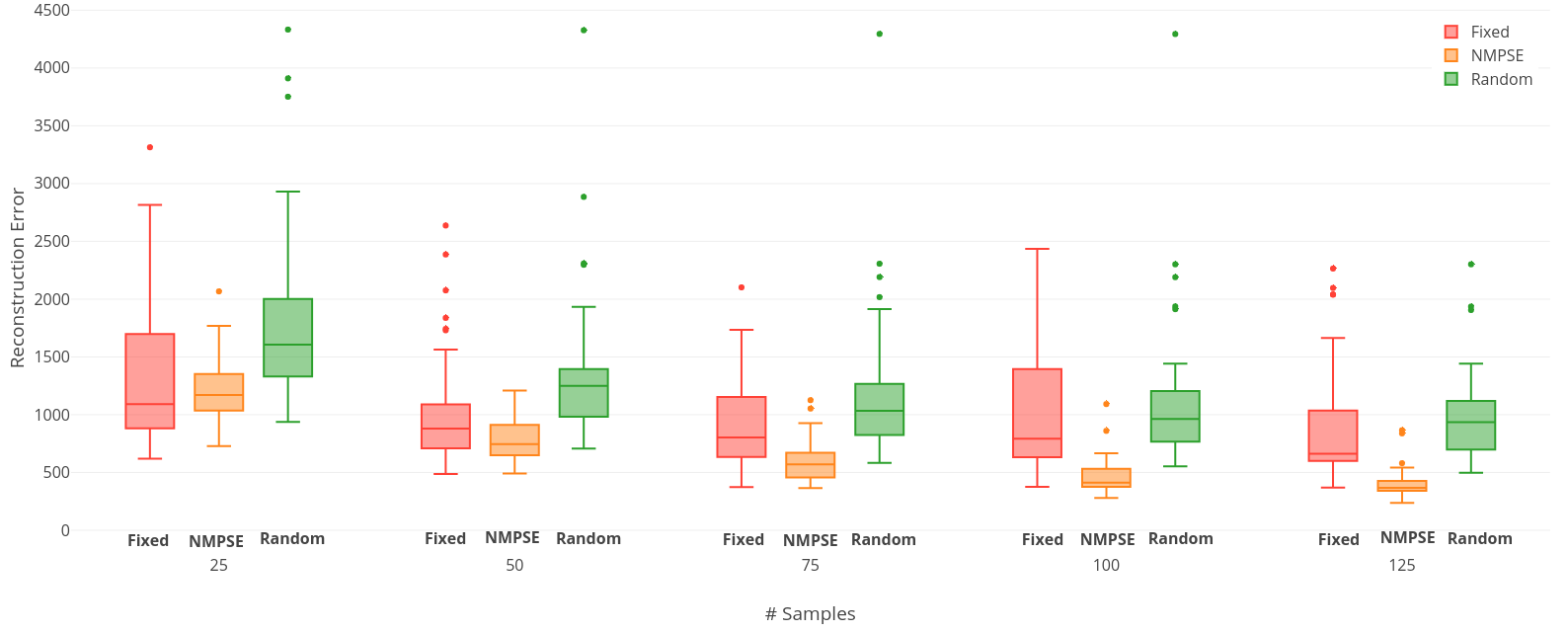}
    \caption{Comparison of the performance of NMPSE (orange) with Fixed Sampling (red) and Random Sampling (green) with varying sampling budgets. All planners were initialized with the same rover configuration.}
    \label{naive}
\end{figure*}
\begin{figure}
    \centering
    \includegraphics[width=8.4cm]{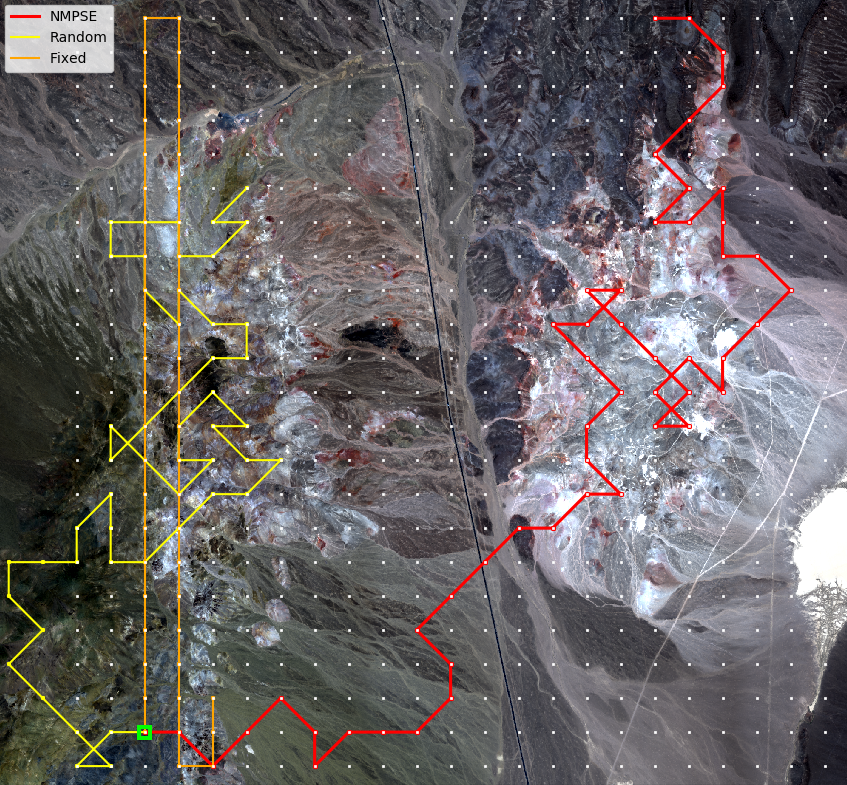}
    \caption{Example paths from NMPSE (red), Random (yellow) and Fixed Step (orange) when constrained by sampling budget overlaid on the map of Cuprite mining site. The white dots on the image refer to discretized grid of sampling points. The green square corresponds to the start location.}
    \label{sampling_path}
\end{figure}
We evaluated NMPSE against two naive planning approaches; namely, random search and fixed step sampling. Figure \ref{naive} displays the boxplots for each of the planners while varying the sampling budget. NMPSE significantly outperformed both random sampling and fixed sampling planners. With increasing sampling budget, NMPSE reduced the reconstruction error at a faster rate than the other two planners. MRE and whether statistical significance ($p < 0.05$) was achieved is displayed in Table \ref{naive}. We evaluated statistical significance with two  one-tailed t-tests: comparing NMPSE with random sampling and fixed sampling. As shown, NMPSE achieved a lower mean with statistical significance in all but one case. We believe NMPSE was not able to achieve $p < 0.05$ in that case due to the sample size being too small. Figure \ref{sampling_path} displays a path obtained from each of the planning strategies from one of the 50 experiments. Observe that NMPSE spends a significant sampling budget in areas with high mineral diversity that lie on the right portion of the image. 
\begin{table}
\caption{Comparison of  Mean Reconstruction Error (MRE) and t-test for NMPSE, Fixed Sampling and Random Sampling planners with different sampling budgets.}
\label{naivetable}
\centering
\begin{tabular}{|l|l|l|l|l|l|}
\hline
Samples & MRE & MRE       & MRE & Fixed & Random   \\ 
          & Fixed      & NMPSE            & Random     & $p < 0.05$ & $p < 0.05$ \\ \hline
25        & 1269.95    & \textbf{1200.79} & 1779.89    & \textbf{N} & Y          \\ \hline
50        & 999.96     & \textbf{783.64}  & 1333.05    & Y          & Y          \\ \hline
75        & 949.32     & \textbf{592.19}  & 1163.42    & Y          & Y          \\ \hline
100       & 976.8      & \textbf{469.61}  & 1090.02    & Y          & Y          \\ \hline
125       & 879.55     & \textbf{396.32}  & 974.09     & Y          & Y          \\ \hline
\end{tabular}
\end{table}
\section{Conclusion}
\label{con}
This work proposed an MDP framework for solving the the SSE problem. By utilizing the inherent structure of the SSE objective, we showed how the problem can be made tractable to apply MDP solvers. We demonstrated the performance of an MCTS based planner against informed and uninformed planning algorithms utilized in this scenario. Our planner reasons about the future states by efficiently evaluating remote sensing measurements with the {\it in situ} measurements. NMPSE achieves lower mean reconstruction error compared to GSS under the same initial and final conditions. Moreover, NMPSE is \textit{scalable} to large orbital images, unlike GSS which must iterate through each pixel of the orbital image for determining a sampling location. 

Having conducted empirical evaluations of the planner using the AVIRIS-NG spectroscopic data, we will evaluate our approach in a field experiment at Cuprite utilizing a rover with a spectrometer on-board. NMPSE will provide waypoints to the robot navigation system which will then generate trajectories between waypoints that satisfy mobility constraints. From a robotics point of view, an interesting area of future research is incorporating safety in navigation, specifically by considering risk in selecting sampling locations to avoid areas that are unsafe for the robot. Current orbital sensors also provide terrain information in the form of Digital Elevation Models (DEM). These DEMs can be used to compute the slope of a region which can inform the planner about safe regions. 
\section*{Acknowledgement}
This research was supported by the National Science Foundation National Robotics Initiative Grant \#IIS-1526667. We acknowledge support of the NASA Earth Science Division for the use of AVIRIS-NG data. We gratefully acknowledge the assistance of Dr. David Thompson at Jet Propulsion Laboratory, California Institute of Technology. U.S. government support acknowledged.
\bibliographystyle{IEEEtran} \bibliography{iros_bibliography}
\end{document}